\documentclass[sigconf,10pt,prologue]{acmart}
\usepackage{adjustbox}
\AtBeginDocument{%
  }

\setcopyright{acmcopyright}
\copyrightyear{2018}
\acmYear{2018}
\acmDOI{XXXXXXX.XXXXXXX}

\acmConference[SEC '24]{The Ninth ACM/IEEE Symposium on Edge Computing}{December 4-7, 2024}{Rome, Italy}
\acmPrice{15.00}
\acmISBN{978-1-4503-XXXX-X/18/06}

\usepackage{array}
\usepackage{booktabs}
\usepackage{multirow}
\usepackage{amsmath}
\usepackage{algorithm}
\usepackage{algpseudocode}
\usepackage{hyperref}
\usepackage{caption}
\usepackage{pbalance}
\usepackage{titlesec}
\titlespacing*{\section}{0pt}{0.1ex}{0.1ex}
\usepackage{colortbl}

\begin{document}

\title{Low-Energy On-Device Personalization for MCUs}



\author{Yushan Huang}
\affiliation{%
  \institution{Imperial College London}
  \city{London}
  \country{UK}}

\author{Ranya Aloufi}
\affiliation{%
  \institution{Imperial College London}
  \city{London}
  \country{UK}}

\author{Xavier Cadet}
\affiliation{%
  \institution{Imperial College London}
  \city{London}
  \country{UK}}

\author{Yuchen Zhao}
\affiliation{%
  \institution{University of York}
  \city{York}
  \country{UK}}

\author{Payam Barnaghi}
\affiliation{%
  \institution{Imperial College London}
  \city{London}
  \country{UK}}

\author{Hamed Haddadi}
\affiliation{%
  \institution{Imperial College London}
  \city{London}
  \country{UK}}



\begin{abstract}
Microcontroller Units (MCUs) are ideal platforms for edge applications due to their low cost and energy consumption, and are widely used in various applications, including personalized machine learning tasks, where customized models can enhance the task adaptation. However, existing approaches for local on-device personalization mostly support simple ML architectures or require complex local pre-training/training, leading to high energy consumption and negating the low-energy advantage of MCUs.

In this paper, we introduce $MicroT$, an efficient and low-energy MCU personalization approach. $MicroT$ includes a robust, general, but tiny feature extractor, developed through self-supervised knowledge distillation, which trains a task-specific head to enable independent on-device personalization with minimal energy and computational requirements. 
$MicroT$ implements an MCU-optimized early-exit inference mechanism called stage-decision to further reduce energy costs. This mechanism allows for user-configurable exit criteria (stage-decision ratio) to adaptively balance energy cost with model performance. We evaluated $MicroT$ using two models, three datasets, and two MCU boards. $MicroT$ outperforms traditional transfer learning ($TTL$) and two SOTA approaches by 2.12 - 11.60\% across two models and three datasets. Targeting widely used energy-aware edge devices, $MicroT$'s on-device training requires no additional complex operations, halving the energy cost compared to SOTA approaches by up to 2.28$\times$ while keeping SRAM usage below 1MB. During local inference, $MicroT$ reduces energy cost by 14.17\% compared to $TTL$ across two boards and two datasets, highlighting its suitability for long-term use on energy-aware resource-constrained MCUs.

\end{abstract}




%
\maketitle

\vspace{-10.5pt}
\section{INTRODUCTION}
\label{sec:intro}


Microcontroller Units (MCUs) are cost-effective and energy-efficient, enabling wide opportunities for Tiny Machine Learning (TinyML)~\cite{dutta2021tinyml}. For example, MCUs can be used in wearables for health monitoring~\cite{haghi2017wearable}, smart home for energy management~\cite{abadade2023comprehensive}, and industrial sensors for predictive maintenance~\cite{kanawaday2017machine}. However, the diverse range of deployment scenarios for MCUs makes them vulnerable to data drift issues.

Data drift~\cite{liang2023comprehensive}, when the distributions of source and target data diverge, can result in a drop in model performance. To solve this challenge, common personalization approaches such as multi-task learning~\cite{zhang2018overview,hu2021unit}, zero-shot learning~\cite{pourpanah2022review}, and cloud-edge methods like federated learning~\cite{mcmahan2017communication,hao2020time,chai2021fedat}, have been proposed. Nevertheless, these approaches have limitations: multi-task learning and zero-shot learning can struggle with the complexity and variability of local data, while cloud-edge strategies rely on frequent and stable communication between the cloud and local resources, often requiring significant computational resources. Therefore, personalizing models locally is essential to maintain their effectiveness.

On-device personalization, performed entirely by the local device without cloud communication, improves efficiency and reduces latency~\cite{goldenberg2021personalization}.
$TinyTL$~\cite{cai2020tinytl} and $SparseUpdate$~\cite{lin2022device} both achieve personalization on MCUs and have been deployed and tested with advanced models. However, our experiments found that despite making significant contributions to model performance, they consume significant energy (detailed in Section~\ref{sec:energy}).
$TinyTL$ adds lite-residual modules to the backbone and only trains these added modules on the MCU. Although these modules have been carefully designed, they still require a significant amount of energy for training.
$SparseUpdate$ only trains the selected layer/channel to improve the efficiency. However, the layer/channel-searching requires access to an entire local dataset, which is not applicable in real scenarios. 
It also requires thousands of computationally intensive search \cite{lin2022device} or pruning \cite{profentzas2022minilearn} processes, resulting in long training times of up to ten days \cite{kwon2023tinytrain}. Moreover, these methods focus solely on local training and overlook local inference, which impacts the energy consumption of the MCU during long-term operation.



\textbf{Challenges.}
Achieving on-device personalization on MCUs presents the following challenges:

(1) MCUs are ultra-resource-constrained, typically with less than 1MB of RAM. This is \~3 orders of magnitude smaller than mobile devices, and \~5-6 orders of magnitude smaller than cloud GPUs \cite{lin2020mcunet}.

(2) MCU personalization requires hardware-software co-design. MCUs do not has operating systems, and model deployment and training needs to be done in $C/C++$. We need to design models in conjunction with existing MCU AI toolkits in order to control memory/energy effectively.

(3) MCUs are energy-aware device. They are widely used in wearable devices and batteryless systems \cite{zhao2022towards,afzal2022battery} given their low cost and energy. Realizing model personalization on MCUs should not compromise their low-energy advantages.

These challenges have led studies to experiment with more capable mobile devices~\cite{wang2022melon,cai2021towards,xu2022mandheling}, like smartphones with 8GB of RAM, unlike MCUs with only 1MB of RAM.

\textbf{Contribution.}
In this study, we introduce $MicroT$, a low-energy model personalization framework for ultra-resource-constrained MCUs \footnote{https://github.com/yushan-huang/MicroT}.
The goal of $MicroT$ is to sustain personalization performance while reducing its energy consumption. We consider that model personalization includes both training and inference and therefore explored model performance and energy consumption in both of these stages.
Our contributions are summarized as follows:

(1) We believe that the transfer learning paradigm is the most energy-efficient approach for the problem of MCU model personalization, as demonstrated by $TinyTL$ \cite{cai2020tinytl} and $SparseUpdate$ \cite{lin2022device}.
However, these two methods place excessive computational burden on the resource-constrained MCU, leading to significant energy consumption.
The $TTL$ paradigm, which involves training only the task head (such as a classifier) locally, remains the most energy-efficient approach and requires the least computational resources.
Therefore, $MicroT$ follows the $TTL$ paradigm, dividing the model into a feature extractor, and a classifier.

(2) $MicroT$ utilizes Self-Supervised Learning (SSL) to obtain a powerful feature extractor.
Although training only the classifier on the MCU can reduce energy consumption and computational demands, the classifier's limited capacity makes it difficult to capture complex features.
We want to avoid placing this computational burden on the MCU to reduce energy consumption.
Thus, naturally, the `feature extractor + classifier' framework led us to consider improving the generality and quality of the feature extractor.
To do so, we explored unsupervised training and ultimately adopted SSL.
Since cloud devices have no access to local data, we utilize a non-local public dataset, which includes a variety of categories and learning samples.
Learning from this public local dataset, we ensure that the feature extractor adequately learns knowledge across a broad range of data.

(3) $MicroT$ integrates SSL with Knowledge Distillation (KD) to reduce the size of the feature extractor. Due to memory constraints, the models deployed on MCUs have limited capacity. We found that directly applying $SSL$ to these small models yields limited performance improvements \cite{azizi2021big} (detailed in Section. \ref{sec:ablation}).
Therefore, we first apply SSL to a large model and generate a dataset based on the embedding features extracted once trained.
Then, a smaller feature extractor learns from both this generative dataset and the original dataset simultaneously.

(4) $MicroT$ uses an MCU-tailored early-exit mechanism to reduce inference energy consumption.
However, existing early exit methods are not suitable for MCUs because they have multiple potential exits and complex exit criteria, which impose significant computational and energy burdens on MCUs.
We propose an MCU-tailored early-exit mechanism that segments the feature extractor into part/full models using a model segmentation fused score considering the trade-off between the performance and efficiency (detailed in Section. \ref{sec: fusedscore}). During MCU inference, $MicroT$ uses classification confidence as the exit criterion on which to perform joint inference, which we call the stage-decision.

(5) To demonstrate the feasibility of $MicroT$, we implement and evaluate it on two models, three datasets, and two MCU boards. $MicroT$ outperforms $TTL$ and two state-of-the-art (SOTA) approaches by 7.01 - 11.60\%, 2.29 - 5.73\%, and 2.12 - 3.68\% in accuracy across two models and three datasets, respectively. During the MCU training phase, $MicroT$ saves 2.03 - 2.28$\times$ energy compared to SOTA methods across two models and two datasets. During the MCU inference phase, $MicroT$ reduces energy consumption by 14.17\% compared to $TTL$ across two models and two datasets.

(6) $MicroT$ provides a configurable stage-decision ratio, allowing users to flexibly choose the balance between performance and energy consumption. It is suitable for a wide range of commercial MCUs and models.

\section{BACKGROUND AND MOTIVATION}
\label{sec:related}

\begin{figure*}[!t]
\begin{center}
\centerline{\includegraphics[width=1\textwidth]{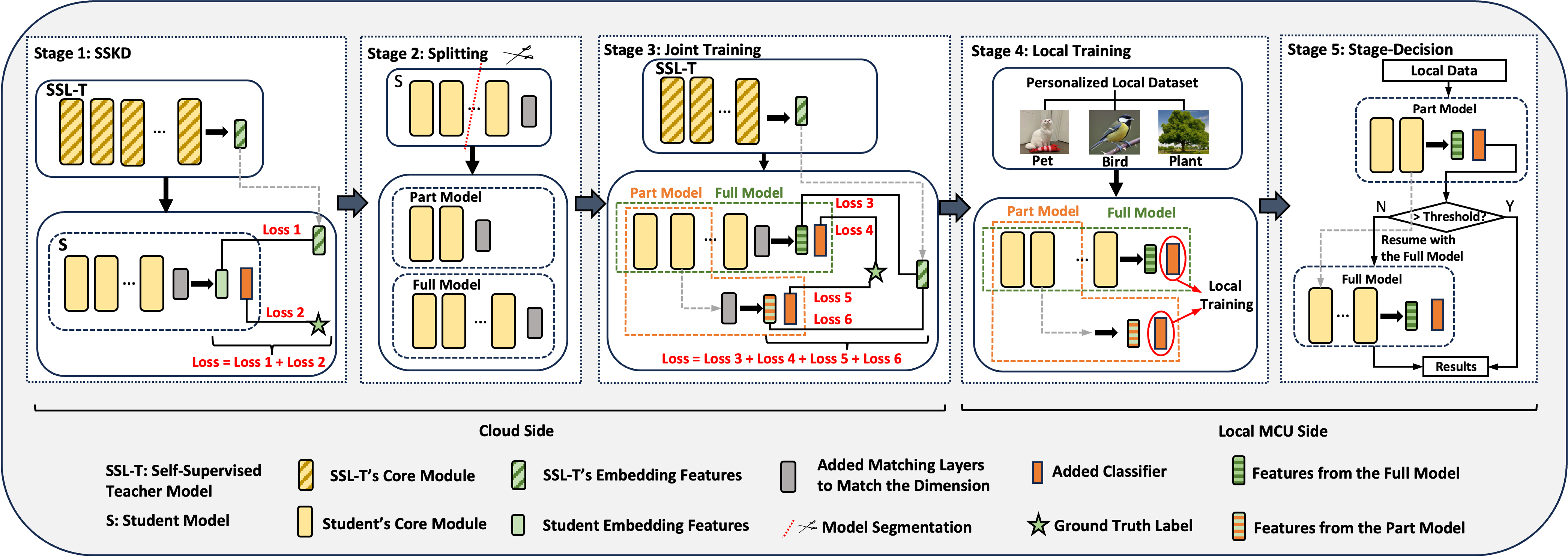}}
\caption{System Overview of MicroT}
\label{fig:system}
\end{center}
\vspace{-20pt}
\end{figure*}

In this section, we provide the necessary background and motivation for $MicroT$.

\textbf{MCU-AI is pervasive.} An important trend in AI deployment is migrating models from cloud devices to edge devices, which can better reduce latency and protect privacy \cite{chen2022mobrecon, chen2022mobile}. MCU-based edge devices are highly suitable for large-scale deployments given their low cost and energy consumption. For instance, MCUs can be used to process signals from microphones, accelerometers, gyroscopes, low-power cameras, and other sensors to enable a range of applications including predictive maintenance \cite{chen2023lopdm}, acoustic detection \cite{mayer2019self}, image detection \cite{afzal2022battery}, and human activity recognition \cite{huang2024analyzing}.

\textbf{MCUs are resource-constrained.} MCUs have limited memory (SRAM) and storage (Flash) budgets. For example, the STM32L4R5ZI is equipped with an ARM Cortex-M4 core and has only 2$MB$ of Flash memory and 640$KB$ of RAM. In comparison, cloud GPUs ($e.g.$, NVIDIA V100 with 16$GB$ memory and TB-PB storage) and mobile devices ($e.g.$, iPhone 11 with 4$GB$ memory and 64GB storage) have \~3 to 6 orders of magnitude more memory and storage.

\textbf{Cloud-edge collaboration personalization.} Federated learning enables a collaborative approach between cloud and local resources, by training a global model in the cloud and fine-tuning locally the complete or partial model to adapt to specific tasks.
However, federated learning relies on frequent and stable communication between the cloud and the local resources. Despite recent efforts to address this issue, federated learning remains challenging to implement in specific scenarios, such as remote mountainous areas. Additionally, this regular communication brings significant energy costs and the risk of privacy leakage. 
Multi-task learning leverages domain information related to the source task contained in the training signals of the target task in order to enhance the model performance.
Multi-task learning relies on the similarity between the different tasks to achieve feature sharing; however, in the absence of prior knowledge on the target task or significant differences exist between the source and target data, multi-task learning can struggle to effectively improve performance.
Zero-shot learning also faces challenges due to its reliance on rich semantic information to enable knowledge transfer across categories. This information can include manually defined attribute vectors, automatically extracted word vectors, context-based embeddings, or combinations of these \cite{pourpanah2022review}.

\textbf{On-device personalization.} Due to the extremely constrained resources of MCUs, some works have proposed leveraging communication between cloud and local devices to enable on-device model personalization. However, frequent communication can result in significant energy consumption, even over the transmission distances in distributed body-wearable devices\cite{gong2023collaborative}. The current on-device training methods, such as $SparseUpdate$\cite{lin2022device} and $TinyTL$\cite{cai2020tinytl}, have their own drawbacks: the former requires the access to an entire local dataset and requires nearly ten days to search for the layers/channels to update and train the model, while the latter requires the inefficient training of numerous additional modules. Although these methods contribute well on model performance, they fall short in considering energy consumption and practical feasibility.

\textbf{Hardware-software co-design.} Unlike cloud and mobile devices, MCUs are bare-metal devices without operating systems and cannot support the advanced programming languages, such as Python, typically used when deploying ML models. The inference and training routines of on-device models instead need to be implemented in $C/C++$. This requires the co-design of models with existing MCU AI libraries and the effective management of memory/storage/energy. Due to a lack of hardware-software co-design, studies often utilize small-resolution datasets ($e.g.$, CIFAR10/100) \cite{liu2023space}, deploy simple ML architectures \cite{zhang2023leveraging}, or only evaluate their method on mobile devices \cite{kwon2023tinytrain,wang2022microcontroller}. 

\textbf{Model inference is also crucial.} Currently, approaches enabling model inference on MCUs mainly focus on model compression techniques such as quantization \cite{xiao2023smoothquant,jacob2018quantization} and pruning \cite{zhu2017prune,liu2020pruning}. However, these methods are static, meaning that once executed on cloud devices and deployed to locally they cannot dynamically adjust the model's inference energy consumption. In the realm of LLMs, early-exit mechanisms \cite{schuster2022confident, wang2023tabi, elhoushi2024layer} are a popular approach to improve inference efficiency. Traditional early-exit mechanisms support multiple potential exit layers and perform early-exiting based on designated exit criteria. However, these methods are difficult to apply on resource-constrained MCUs, because supporting multiple potential exit layers and complex/frequent exit criteria consumes significant MCU resources.

\section{DESIGN}
\label{sec:system}

In this section, we introduce the system overview of $MicroT$. Then, we separately present the methods involved in $MicroT$ from both the cloud side and the local MCU side.

\vspace{-10pt}
\subsection{Overview}
\label{sec:overview}

$MicroT$ consists primarily of five core components, as shown in Fig. ~\ref{fig:system}. On the Cloud side, $MicroT$ includes self-supervised knowledge distillation (SSKD), model splitting, and joint training. On the Local MCU side, $MicroT$ includes local training and stage-decision. Next, we undertake an illustration of $MicroT$'s characteristics, including an overall description of these five core components of $MicroT$.

\textbf{The design philosophy of MicroT is based on transfer learning.} 
In the context of MCU personalization, transfer learning exhibits unique advantages: (1) it does not require frequent communication with cloud devices, and (2) it does not require cloud devices to have extensive prior knowledge of local data, as it can adapt to local personalized data by training/fine-tuning. $TinyTL$ and $SparseUpdate$ are based on transfer learning; both require a pre-trained source model/backbone and perform local training or fine-tuning. 



\textbf{Feature extractor + classifier paradigm.} $MicroT$ adopts the $TTL$ approach, dividing the model into a feature extractor and a classifier, guided by energy and computational requirements. We found that the performance improvements obtained by existing MCU $TTL$ approaches, such as $TinyTL$ and $SparseUpdate$, are at the cost of increased energy and computational demands (detailed in Section. \ref{sec:energy}), undermining the advantages of MCUs. While the `feature extractor + classifier' paradigm results in inferior model performance, it offers the lowest energy consumption and computational demands. We believe that, from a practical standpoint, the most direct approach for optimization is to retain this paradigm and optimize its performance.

\textbf{MicroT supports a powerful but tiny feature extractor.} The inferior model performance of the `feature extractor + classifier' paradigm is mainly due to limited capacity of the classifier, making it difficult to capture complex feature variations. Since the classifier is trained on the MCU, we believe it is more practical to improve the feature extractor's ability to capture complex feature variations, as the feature extractor is trained on a cloud device and can fully utilize its computational resources. The design goal of this feature extractor is to be sufficiently generalizable and versatile, thereby helping the classifier achieve stable model performance on a wide range of personalized local data. To enhance the versatility of the feature extractor, we adopted self-supervised learning ($SSL$), which can improve a model's versatility through strategies such as data augmentation. Additionally, since cloud devices do not have access to local data, we utilize a non-local public dataset in the cloud which includes a variety of categories and learning samples ($e.g.$, ImageNet). This ensures that the feature extractor can learn adequately broad knowledge across a range of data.

Given the limited memory of MCUs, we can only deploy lightweight models, such as MCUNet \cite{lin2020mcunet}. In our experiments, we found that directly applying $SSL$ to these lightweight models resulted in limited performance improvements, with the model capacity causing a bottleneck \cite{azizi2021big} (detailed in Section \ref{sec:ablation}). Therefore, to optimize the feature extractor, we combined SSL and KD (SSKD), as shown in Fig. \ref{sec:overview} Stage 1. Specifically, we first use SSL to train a large model on a public dataset, then use the large model to obtain a generative dataset, and finally, train a smaller feature extractor on the generative dataset and the public dataset.


\textbf{MicroT offers low energy costs for model training on local devices.} When the MCU receives the feature extractor, it initializes the classifier based on the local task and then trains it locally on the device using a local dataset, as shown in Fig. ~\ref{fig:system} Stage 4. Thanks to the powerful but tiny feature extractor, the classifier can achieve improved performance with only simple task head ($e.g.$, a classifier), and maintain stability even when faced with unseen or new local data. This is consistent with our expectations that the `feature extractor + classifier' paradigm would result in lower computational demands and energy consumption.


\textbf{MicroT offers low system costs for model inference on local devices.} The system cost of model inference will determine the long-term or even life-long energy cost of the MCU. Inspired by the early-exit mechanism, we find that not all data requires full model inference. Even using only the front portion of the model can yield results consistent with full inference,  reducing the system consumption of the MCU. Traditional early-exit mechanisms set  potential early-exit points at several layers of the model. We do not adopt this approach because frequent exit criteria decisions and multiple classifier heads increase the system consumption of the MCU. Instead, we split the model into two: the part model and the full model, as shown in Fig. ~\ref{fig:system} Stage 2. The part model is a subset of the full model, reducing the memory footprint when deploying part and full models. We propose a model split fused score for model segmentation, which considers both the performance and system consumption of the model to efficiently identifies the optimal model partitioning point. To ensure alignment between the two models, we also employ joint training, as shown in Fig. ~\ref{fig:system} Stage 3. Finally, the part and full models work collaboratively on the local MCU, a process we refer to as stage-decision, as shown in Fig.~\ref{fig:system} Stage 5. The data is initially analyzed by the part model, followed by selectively activating and awakening the full model based on a confidence threshold. To determine the confidence threshold, we propose a method based on median-quartile values that can efficiently decide a local confidence threshold with just five local data points, as demonstrated experimentally (detailed in Section. \ref{sec:sample}).

\textbf{MicroT is flexible and adaptable.} $MicroT$ does not require specific platform support and is applicable across commercial MCU platforms such as the STM32 series and Espressif ESP series. $MicroT$ is also compatible with various models, as long as the model and operators are supported by the MCU platform and meet the resource requirements. In the local stage-decision phase, $MicroT$ enables users to flexibly balance model performance and energy cost by adjusting the stage-decision ratio, sacrificing model performance to reduce energy cost or energy cost to improve performance.

\vspace{-10pt}
\subsection{Self-Supervised Knowledge Distillation}
\label{sec:sskd}
As illustrated in Section~\ref{sec:overview}, to reduce the system costs and memory requirements for local model training on MCUs, we divide the model into a feature extractor and a classifier. The design of the feature extractor had two considerations: (i) The necessity for the extracted features to be generalizable towards the personalized local data, enabling the added local task head to be trained at low system cost; (ii) The unavailability of local data in the cloud for training. To address these considerations, we utilize SSL to improve the generalizability of the feature extractor, and then utilize KD to obtain a powerful but tiny feature extractor.


Firstly, we apply SSL to the teacher. The rationale for not applying SSL directly to the student model is that $SSL$ generally offers greater performance improvements for larger models. Given the simpler architecture of the student model, direct SSL application may only offer limited improvements \cite{azizi2021big} (details in Section. \ref{sec:ablation}). To obtain a general teacher model, we follow the paradigm of SSL \cite{oquab2023dinov2}, creating two models: one as the guide model and the other as the exploration model, to support unsupervised feature learning. These models process different variations of the same image via a series of data augmentation techniques, including random cropping, color perturbation, and flipping. The guide model is designed to use an exponential moving average (EMA) approach to smooth its parameter updates. In contrast, the exploration model updates its parameters directly through back-propagation, attempting to mimic the guide model's outputs by learning features directly from the varied perspectives. The loss function is:

\begin{equation}
L_{\text{SSL}} = - \sum_{i} \text{softmax}\left(\frac{g_i}{\tau_g}\right) \log \text{softmax}\left(\frac{e_i}{\tau_e}\right)
\end{equation}

where $g_i$ and $e_i$ represent the output logits of the guide model and the exploration model. $\tau_g$ and $\tau_e$ are temperature parameters that adjust the sensitivity of the $softmax$. This setup encourages the exploration model to learn the behavior of the guide model by processing different data views, while the guide model's EMA strategy ensures it gradually adapts and improves the data representation capturing ability.

After obtaining the guide model trained through SSL, this model transitions into the teacher model for the KD phase. During KD, to enable the source model to better learn from the teacher model, we extract the embedding features of the teacher model. Specifically, we pass the public dataset through the trained teacher model and extract the embedding outputs from the layers before the head (feature extractor) to obtain a generative dataset. To ensure that the dimensions of the source model's feature extractor $P$ match the dimensions of the teacher model $Q$, we add a fully connected layer (Matching Layer) after the source model's feature extractor to align the output dimensions $(P, Q)$ \cite{peplinski2020frill}. The distillation loss can be formulated as follows:

\begin{equation}
L_{\text{distill}} = \sum_{j} [\alpha * L_{\text{MSE}}(p_{j}, q_{j}) + L_{\text{CE}_{j}}]
\end{equation}

where $\alpha$ is the weight, $p$ is the extracted features from the source model's feature extractor, $q$ is the extracted features from the teacher model's feature extractor, $L_{\text{MSE}}$ is the Mean Squared Error, and $L_{\text{CE}}$ is the Cross Entropy Loss between the source model's final output and the ground truth labels.

We utilize a non-local public dataset in the cloud that requires a variety of categories and learning samples to maximize the generalizability of the teacher and feature extractor (given the lack of prior knowledge of local data in the cloud).



\begin{algorithm}[!t]
\footnotesize
\caption{Determine Optimal Model Segmentation Point}
\begin{algorithmic}[1]
\State Initialize:
\State $PM \gets \text{Split full model into part model based on module}$
\State $A_{full} \gets \text{Compute accuracy of the full model}$
\State $M_{full} \gets \text{Compute MAC of the full model}$
\State $F_{max} \gets 0$
\State $OptimalSplit \gets 1$
\For{$i = 1$ to $\text{length of } PM$}
    \State $A_i \gets \text{Compute accuracy of } PM[i]$
    \State $M_i \gets \text{Compute MAC of } PM[i]$
    \If{$i > 1$}
        \State $\Delta A \gets A_i - A_{i-1}$
        \State $\Delta M \gets M_i - M_{i-1}$
        \State $G \gets \text{Normalize}(\Delta A / \Delta M)$
    \EndIf
    \State $R_A \gets (A_{full} - A_i) / A_{full}$
    \State $R_A^{norm} \gets \text{Normalize}(R_A)$
    \State $R_M \gets (M_{full} - M_i) / M_{full}$
    \State $R_M^{norm} \gets \text{Normalize}(R_M)$
    \State $F_{score} \gets [3 * (1 - R_A^{norm}) * G * R_M^{norm}] / [(1 - R_A^{norm}) + G + R_M^{norm}]$
    \If{$F_{score} > F_{max}$}
        \State $F_{max} \gets F_{score}$
        \State $OptimalSplit \gets i$
    \EndIf
\EndFor
\State \Return $OptimalSplit$
\end{algorithmic}
\label{alg:split}
\end{algorithm}

\subsection{Model Segmentation and Joint Training}
\label{sec: fusedscore}

To reduce the energy cost during local inference on MCUs, $MicroT$ uses model segmentation and joint training. It segments the feature extractor into a part model and a full model, enabling stage-decision and joint inference on the MCU. To rationally split the feature extractor, we propose a model split fused score $F_{\text {score}}$, then adopt joint training to further optimize the performance of the part/full models. Considering the absence of local data in the cloud, we utilize the non-local public dataset to determine the optimal segmentation point. The definition of $F_{\text {score}}$ is:

\begin{equation}
F_{\text{score}} = \frac{3 \times (1 - R_{A}^{\text{norm}}) \times G \times R_{M}^{\text{norm}}}
{(1 - R_{A}^{\text{norm}}) + G + R_{M}^{\text{norm}}}
\label{eq:fscore}
\end{equation}

where $F_{\text{score}}$ is the score used to evaluate the segmentation point. $R_{A}^{\text{norm}}$ represents the normalized ratio of accuracy loss, which measures the decrease in the part model's accuracy relative to the full model's accuracy. $R_{M}^{\text{norm}}$ is the normalized Multiply–Accumulate ($MAC$) reduction ratio, which quantifies the decrease in the number of $MACs$ in the part model's relative to the number of $MACs$ in the full model. $G$ is the normalized gain ratio, representing the trade-off between the increased accuracy and $MACs$ from the part model.

In determining the optimal model segmentation point, we consider both model performance and computational efficiency, acknowledging the impact of computational demand on energy cost \cite{he2018amc,tan2019mnasnet}. We use accuracy as the performance indicator for the model and the number of $MACs$ as the indicator for computational demand. Specifically, as detailed in Algorithm ~\ref{alg:split}, we initially split the complete model into part models based on its modules, and then compute the accuracy ($A$) and $MACs$ ($M$) for each part model. Subsequently, Algorithm ~\ref{alg:split} calculates the accuracy improvement ($\Delta A$) and change in the number of $MACs$ ($\Delta M$) for each part model relative to the previous part model, and from this calculates the normalized gain ratio ($G$). Additionally, Algorithm ~\ref{alg:split} calculate the accuracy reduction rate ($R_A$) and $MAC$ reduction rate ($R_M$) for each part model relative to the complete model and normalizes these ratios. By integrating these metrics, we calculate a composite score ($F_{\text {score}}$) for each part model. After calculating $F_{\text {score}}$ for all part models, we select the split point with the highest composite score as the optimal split point, and obtains the part model, to be embedded, and full model, as shown in Fig. ~\ref{fig:system} Stage 2. Unlike traditional early-exit mechanisms, we omit multiple potential exit layers and efficiently split the model into two embedded parts based on model performance and computational demand, thereby avoiding redundant exit criterion calculations.

Following the model segmentation, we perform joint training on top of the pre-trained feature extractor obtained from the SSKD, as shown in Fig. ~\ref{fig:system} Stage 3. We add a matching layer for the part model to align with the feature dimensions of the teacher. Subsequently, we further fine-tune the part and full model with a composite loss function $L_{\text{JT}}$:

\begin{equation}
L_{\text{JT}} = \sum_{k} [\alpha * L_{\text{MSE}}(p_{k}^{\text{F}}, q_{k}) + \beta * L_{\text{MSE}}(p_{k}^{\text{P}}, q_{k}) + L_{\text{CE}{j}}^{\text{F}} + L_{\text{CE}{j}}^{\text{P}}]
\end{equation}

where $\alpha$ and $\beta$ are weights, $p_{F}$ is the extracted features from the full model, $p_{P}$ is the extracted features from the part model, $q$ is the extracted features from the teacher model, $L_{\text{MSE}}$ is the Mean Squared Error, $L_{\text{CE}{j}}^{\text{F}}$ is the Cross Entropy Loss between the full model's final output and the ground truth labels, $L_{\text{CE}{j}}^{\text{P}}$ is the Cross Entropy Loss between the part model's final output and the ground truth labels. Since both the part and full models learn the teacher model's embedding features, their alignment is ensured.

\vspace{-5pt}
\subsection{Classifier Training}
\label{sec:training}
The cloud-trained feature extractor (with \texttt{INT8} quantization and only \~0.5\% accuracy loss, detailed in Section. \ref{sec:ablation}) is deployed on the MCU, which locally builds and trains classifiers to process local personalized data, as shown in Fig. ~\ref{fig:system} Stage 4. 
The general features learned by the feature extractor enable the use of a simple classifier, which can achieve sufficient model performance while reducing memory constraints on the MCU. 
Unlike cloud and mobile devices, MCUs do not have an operating system and cannot support high-level programming languages such as Python for deploying ML models. Therefore, the deployment, inference, and training of classifiers on local MCUs are all carried out using C/C++. We develop code on AI toolboxes from commercial MCUs, such as STM32 X-CUBE-AI. Building even a single-layer classifier requires numerous key steps:
1) Defining the classifier's architecture, including layer count, the neurons per layer, input and desired output of the Neural Networks, layers (excluding the input layer), and the learning rate; 2) Structuring parameter matrices, encompassing weight matrix, bias array, output array, and error (the partial derivative of the total error with respect to the weighted sum); 3) Defining activation functions; 4) Implementing a function to load datasets and models; 5) Creating a model construction function for memory allocation and model establishment; 6) Developing a training function, which includes forward and backward propagation processes; 7) Instituting a function to save the model; 8) Developing a function to release memory.

To alleviate memory during classifier training, the MCU offloads the memory used by the feature extractor prior to training, retaining only its extracted features. Given the MCU's limited memory capacity, the batch size is set to one. We utilize a stage-training approach to train the part/full model classifiers collaboratively in order to save energy. Specifically, when an input passes through the model, it first goes through the part model, which is used to train the part model's classifier. We save the intermediate values from the layer preceding the part model classifier and use these values to continue inference and train the full model's classifier.

Since local data acquisition may not be continuous, we also consider the issue of data storage to facilitate classifier training. $MicroT$ does not need to store the raw data, only the features extracted from the raw data by the feature extractor; after the raw data passes through the feature extractor, an embedding feature with a dimension of 512 is generated. Storing this feature in \texttt{FLOAT32} format in Flash memory occupies only 2KB. This is acceptable for the Flash memory of commercial MCUs. For example, deploying MCUNet \cite{lin2020mcunet} on the STM32L4R5ZI (with 2MB Flash memory), requires \~0.91MB. In this case, there will still be \~1.09MB of Flash memory available, which could store another \~500+ local samples.

\begin{algorithm}[!t]
\footnotesize
\caption{Stage-Decision}
\begin{algorithmic}[1]
\State Initialize:
\State $S_{train} \gets \text{Set of N training samples to determine threshold}$
\State $S_{infer} \gets \text{Set of M new samples for inference}$
\State $PartModel \gets \text{Function to process samples using the part model}$
\State $FullModel \gets \text{Function to process samples using the full model}$
\State $C \gets []$ \text{Array to store confidence scores}
\State $R \gets []$ \text{Array to store results after stage-decision}
\State $AdjustFactor \gets$ \text{Default = 1, User-defined factor to adjust the threshold}

\State Determine the median confidence threshold from training samples:
\For{$i = 1$ to $N$}
    \State $features_{ip}, c_{ip} \gets PartModel(S_{train}[i])$
    \State Append $c_{ip}$ to $C$
\EndFor
\State Sort $C$ in ascending order
\If{$N$ mod $2$ = 1}
    \State $C_{median} \gets C[\frac{N+1}{2}]$
\Else
    \State $C_{median} \gets \frac{C[\frac{N}{2}] + C[\frac{N}{2} + 1]}{2}$
\EndIf
\State $Threshold \gets C_{median} \times AdjustFactor$

\State Perform stage-decision on inference samples:
\For{$i = 1$ to $M$}
    \State $result_{ip}, c_{ip} \gets PartModel(S_{infer}[i])$
    \If{$c_{ip} < Threshold$}
        \State $features_{ip} \gets \text{Features before output layer of the } PartModel$
        \State $result_{if} \gets FullModel(features_{ip})$
        \State $R[i] \gets result_{if}$
    \Else
        \State $R[i] \gets result_{ip}$
    \EndIf
\EndFor

\State \Return $R$
\end{algorithmic}
\label{alg:stage_decision}
\end{algorithm}

\vspace{-10pt}
\subsection{Stage-Decision}
\label{sec:stage}

Upon training completion, $MicroT$ implements stage-decision to further reduce energy cost during inference, as shown in Fig. \ref{fig:system} Stage 5. Stage-decision is an early-exit mechanism which involves processing inputs by the part model and calculating confidence scores. If the score falls below a set threshold, the full model will resume and continue inference from the stage preceding the part model's output layer. Note that, in local training and stage-decision, data preprocessing is necessary, including resizing and normalization of inputs, which should match the preprocessing steps in the cloud.

Setting an appropriate threshold is crucial. We determine the threshold using a median-quantile method based on the confidence score sequence of the part model. The quantile value of the sequence can determine the ratio of input samples (e.g. 0.5 using the median) needing to be processed only by the part model without further activating the full model. This ratio is named the stage-decision ratio, and can be adjusted by the $AdjustFactor$ function, as shown in Algorithm \ref{alg:stage_decision}. The $AdjustFactor$ function allows users to balance model performance with energy cost according to their specific needs. By raising the $AdjustFactor$, and consequently decreasing the stage-decision ratio, users can improve model performance. Accordingly, lowering the $AdjustFactor$ increases the ratio, leading to reduced energy consumption. 

Our experiments (detailed in Section~\ref{sec:sample}) show that the median and threshold can be accurately determined with just a few samples, thereby avoiding excessive resource consumption on the MCU. We define the median confidence score, where 0.5 of the input samples are processed only by the part model, as the `standard stage-decision ratio'.




\section{IMPLEMENTATION}
\label{sec:implementation}

In this section, we introduce our implementation, models, datasets, performance metrics, and baselines.

\vspace{-5pt}
\subsection{MicroT Prototype}
\label{sec: prototype}
\begin{figure}[!t]
\begin{center}
\centerline{\includegraphics[width=0.3\textwidth]{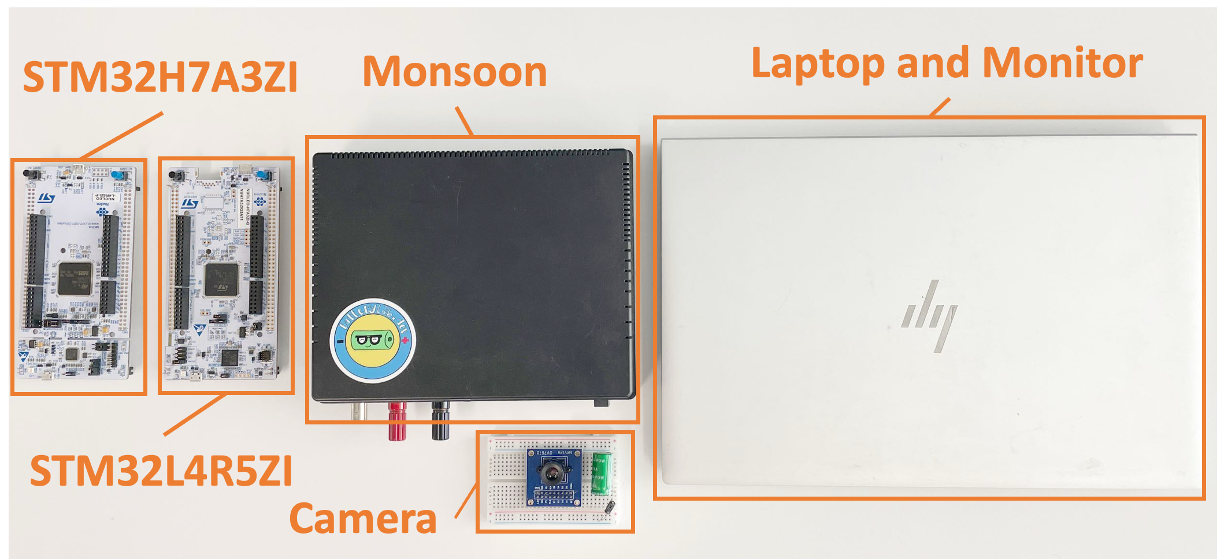}}
\caption{MicroT Prototype Overview}
\label{fig:photo}
\end{center}
\vspace{-20pt}
\end{figure}

We conduct experiments on the STM32 MCUs, as they are representative of mature commercial MCUs. We utilize STM32's ML tool X-CUBE-AI \cite{st_xcubeai} to deploy $MicroT$. X-CUBE-AI contains all the basic functions required for deploying ML on commercial MCUs, including model inference and \texttt{INT8} quantization support. Based on X-CUBE-AI, we modify its generated $C/C++$ code to implement data preprocessing, model training and inference, threshold adjustment, and stage-decision. We run $MicroT$ on the STM32H7A3ZI and STM32L4R5ZI boards specifically, as these represent high-performance and low-power MCUs, respectively. The description of these two MCU boards is shown in Table. \ref{tab:board}.
We utilize the Monsoon High Voltage Power Monitor \cite{msoon} to set the MCU power supply voltage to 1.9$V$ \cite{chen2022underwater}. 
We set the MCU board frequency to 120MHz for the on-device inference and training. Fig.~\ref{fig:photo} shows an overview of the $MicroT$ prototype. A camera is essential in the complete system, but to simplify our experiments, we simulate image acquisition through serial communication.

\subsection{Models and Datasets}
\label{sec: dataset}
\begin{figure}[!t]
\begin{center}
\centerline{\includegraphics[width=0.4\textwidth]{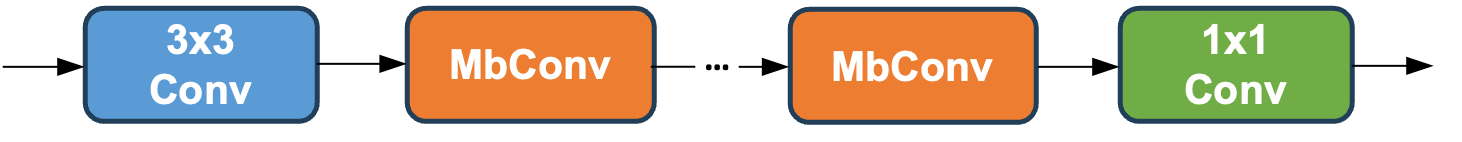}}
\caption{Main Structure of ProxylessNAS and MCUNet}
\label{fig:mbconv}
\end{center}
\vspace{-20pt}
\end{figure}

We select a ViT-based model architecture, named ViT-S \cite{oquab2023dinov2}, as the teacher model. For the student models, to be deployed on the MCUs, we utilize ProxylessNAS\_w0.3 \cite{cai2018proxylessnas}, as well as MCUNet\_int3 \cite{lin2020mcunet}. These are commonly used lightweight CNN and module-based models that balance accuracy, memory footprint, and energy consumption \cite{sadiq2023enabling, sun2022entropy}. For MCUNet and ProxylessNAS, both architectures are fundamentally characterized by Mobile Inverted Bottleneck Convolutions (MBConv), as shown in Fig.~\ref{fig:mbconv}. 

The datasets are categorized into cloud datasets and local datasets. For the cloud datasets, we utilize ImageNet for SSL teacher training, SSKD, model splitting, and joint training. ImageNet provides sufficiently rich samples for the feature extractor to learn from the teacher. For the local datasets, we utilize several datasets to represent the personalized local data, including The Oxford-IIIT Pet (Pet) \cite{petdataset}, CUB-200-2011 Bird (Bird) \cite{WahCUB_200_2011}, and PlantCLEF 2017 (Plant) \cite{goeau2017plant}. These three datasets do not overlap with ImageNet and have distinct data distributions. We use 224$\times$224 as our input resolution.

The Pet dataset \cite{petdataset} contains 3,680 images and 37 distinct pet species. The bird dataset \cite{WahCUB_200_2011} contains a collection of 11,788 images in 200 distinct categories. For the Plant dataset \cite{WahCUB_200_2011}, we select a subset based on image quantities per category, containing 20 plant categories with a total of 11,660 training images. We randomly split the training/testing/validation datasets by the ratio of 8:1:1.

On the cloud side, we perform SSKD from scratch on the feature extractor for 50 epochs with a learning rate of 0.01 and the SGD optimizer. Then we split the pre-trained feature extractor into part/full models, and further perform joint training for 10 epochs with a learning rate of 0.005 and the SGD optimizer. Note that, to implement SSKD and joint training, we add a matching layer before the head of the part/full models to match the dimension of the teacher model's embedding features. This matching layer is removed after training, leaving only the original feature extractor of the part/full models. On the local MCU side, we deploy the trained part/full models, and add a single-layer task head for both. Due to memory limitations, the batch size is set to one, We use the SGD optimizer and a learning rate of 0.1 \cite{lin2022device}.

\begin{table}[t]
\footnotesize
\caption{The experimental MCU boards}
\begin{tabular}{ccccc}
\hline
\textbf{MCU Board} & \textbf{Processor} & \textbf{Flash} & \textbf{RAM} & \textbf{Max Frequency} \\ \hline
STM32H7A3ZI        & Arm Cortex-M7      & 2MB            & 1.18MB       & 280MHz                 \\
STM32L4R5ZI        & Arm Cortex-M4      & 2MB            & 640KB        & 120Mhz                 \\ \hline
\end{tabular}
\label{tab:board}
\end{table}

\subsection{Performance Metrics}
\label{sec: metrics}


\textbf{Model Performance.} We utilize accuracy to assess model performance on personalized local data. 

\textbf{System Cost.} We mainly focus on the energy consumption of local training and inference on MCUs. We measure and calculate the energy consumption using the Monsoon High Voltage Power Monitor \cite{msoon} with 50$Hz$ sampling rate. Specifically, we utilize Monsoon \cite{msoon} to set the input voltage ($U$) to 1.9V \cite{chen2022underwater} and measure the time ($t$) and average current ($I$). Then we calculate the average power ($P = UI$) and average energy consumption ($E = Pt$). To obtain stable values, we only utilize the current recorded after running for 2 $min$. 
All results are an average of ten repeat experiments.

\vspace{-5pt}
\subsection{Baselines}
\label{sec:baseline}
For model performance, we compare $MicroT$ with four baselines: (1) $None$ does not perform any on-device training, and the local classifier is randomly initialized; (2) $TTL$ is traditional transfer learning, which only trains the task head \cite{wu2020emo}; (3) $TinyTL$ adds lite-residual modules to the model backbone and freezes the backbone during training \cite{cai2020tinytl}; (4) $SparseUpdate$ statically selects the layers and channels to update by multiple gradient computation, and then trains these layers/channels online \cite{lin2022device}. To demonstrate the improvements brought by $MicroT$, we use the same feature extractor (MCUNet and ProxylessNAS) for $MicroT$ and all baselines. We also use the same number of training epochs for all methods and grid search to determine the optimal learning rate. To test the optimal performance of the methods, we utilized the complete dataset and split it into training/testing/validation dataset with a ratio of 8:1:1. Although $MicroT$'s stage-decision ratio is configurable, for easier comparison with the baseline(s), we utilize the standard stage-decision ratio of 0.5.

For the system cost of local training, we selected $TinyTL$, $TTL$, and $SparseUpdate$ as the baselines. However, we found that $SparseUpdate$ layer/channel searching requires access to the entire local dataset and thousands of computationally intensive search \cite{lin2022device} or pruning \cite{profentzas2022minilearn} processes, resulting in a training time of up to ten days \cite{kwon2023tinytrain}.
Additionally, because $SparseUpdate$ is static, each time a new task head needs to be added, this complex search process must be repeated. Therefore, we only compared $MicroT$ with $TinyTL$ and $TTL$.

For the system cost of local inference, we compare $MicroT$ with $TTL$ ($i.e.$, traditional `full model' inference).


\section{MODEL PERFORMANCE}
\label{sec:performance}

\begin{table}[!t]
\caption{Results of ML performance with varying batch sizes.}
\footnotesize
\begin{tabular}{ccc}
\hline
Batch Size         & Feature Extractor       & Accuracy \\ \hline
\multirow{2}{*}{128 (GPU)} & MCUNet  & 80.24  \\ \cline{2-3} 
                                    & ProxylessNAS   & 77.45  \\ \hline
\multirow{2}{*}{1 (MicroT)}         & MCUNet  & 79.31  \\ \cline{2-3} 
                                    & ProxylessNAS   & 76.28  \\ \hline
\end{tabular}
\label{tab:match}
\vspace{-10pt}
\end{table}

In this section, we present our evaluation of $MicroT$, with the goal of answering a set of key questions.

\subsection{Does MicroT's Local Training Match Cloud-Level Accuracy?}
\label{sec:matchacc}
Due to memory limitations, when training our classifier on the MCUs, the batch size is set to one. However, this reduces the efficiency of the experiments, as it does not utilize hardware parallelism. Therefore, at the start of the experiments, we investigate the performance gap between classifier training on MCUs with a single-batch and on the GPU with a batch size of 128, to understand if the latter can be used as an approximation of the former to improve the efficiency of the experiments. The training settings on the MCU and cloud are kept consistent, as described in Section. \ref{sec: dataset}. The reason we don't use momentum is that for a single-batch setup, momentum does not benefit model performance and can in fact increase memory usage \cite{lin2022device}. The classifier training on the MCUs is based on $C/C++$, whereas on the GPU is based on $Python$. Due to the low efficiency of single-batch training, we randomly select 40 images from each class in the Pet dataset (a total of 1480 images). 

Our results are shown in Table. ~\ref{tab:match}. We find that single-batch training on the MCU results in similar accuracy to training on the GPU with a 128 batch. This enables us to use the results of batch training on the GPU for evaluation. Therefore, unless otherwise stated, we default to reporting the classifier training results on the GPU, based on the aforementioned classifier training set up.

\begin{figure}[!t]
\begin{center}
\centerline{\includegraphics[width=0.5\textwidth]{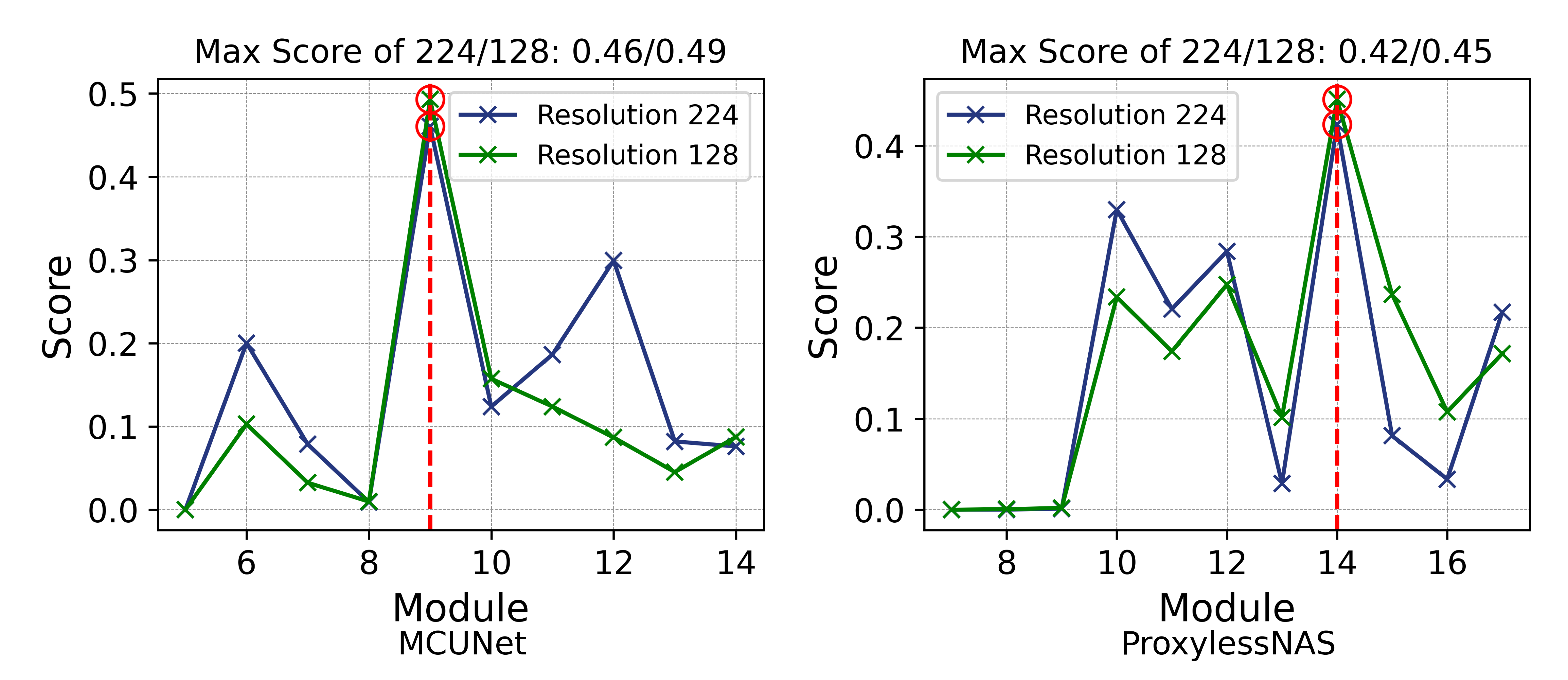}}
\caption{The model segmentation fused scores for MCUNet and ProxylessNAS. The X-axis represents the part model up to certain module, and Y-axis represents the fused score.}
\label{fig:SplitScore}
\end{center}
\vspace{-25pt}
\end{figure}

\subsection{How Does the Performance of MicroT Compare to the Baselines?}
\label{sec:stateofart}
In these experiments, we compare the performance of $MicroT$ with our baselines mentioned in Section \ref{sec:baseline}. We first introduce the deployment process of $MicroT$, and finally compare the model performance.

Once the SSL teacher model is trained and SSKD is applied to the feature extractor, we perform model segmentation. The optimal model segmentation point is determined by the model segmentation fused score (as detailed in Section. ~\ref{sec: fusedscore}), 
Note that we do not want the part model to be too small such that its performance is limited nor do we want it to be too close to the full model, as this would result in minimal optimization of the system cost. Therefore, for MCUNet, we only analyze modules 4 to 14, and for ProxylessNAS, we analyze modules 6 to 17. The model segmentation fused scores of MCUNet and ProxylessNAS are shown in Fig. ~\ref{fig:SplitScore}.

We observe that the optimal segmentation point, based on Algorithm \ref{alg:split}, for MCUNet is at the 9th module, while for ProxylessNAS it is at the 14th module. In these two models, the number of MACs generally increases steadily with the depth of the network. However, at the optimal model segmentation points, there is a significant improvement in accuracy, indicating that the model has extracted key features at this depth and structure. We believe this depth offers the best cost-performance trade-off. We then perform model segmentation evaluation at 128 resolution, and find that the optimal model segmentation points are at the same depths for the same models. This suggests that for a given model, the optimal segmentation point may have a certain stability, and the main factor influencing this point is the model architecture.

\begin{table}[!t]
\caption{Top-1 accuracy results of MicroT and the baselines. MicroT achieves the highest accuracy with two DNN architectures on three cross-domain datasets.}
\footnotesize
\begin{tabular}{ccccc
>{\columncolor[HTML]{ECF4FF}}c }
\hline
\textbf{Model} &
  \textbf{Method} &
  \textbf{Pet} &
  \textbf{Bird} &
  \textbf{Plant} &
  \textbf{Avg.} \\ \hline
 & None         & 42.53          & 20.14 & 35.31          & 32.66 \\
 & TTL          & 74.20          & 43.84 & 47.01          & 55.02 \\
 & TinyTL       & 76.21          & 47.12 & \textbf{53.81} & 59.05 \\
 & SparseUpdate & 79.09          & 49.77 & 50.73          & 59.86 \\ \cline{2-6} 
\multirow{-5}{*}{MCUNet} &
  \cellcolor[HTML]{FFF2DF}MicroT (Ours) &
  \cellcolor[HTML]{FFF2DF}\textbf{81.21} &
  \cellcolor[HTML]{FFF2DF}\textbf{52.85} &
  \cellcolor[HTML]{FFF2DF}52.92 &
  \cellcolor[HTML]{FFF2DF}\textbf{62.33} \\ \hline
 & None         & 37.41          & 19.44 & 33.97          & 30.27 \\
 & TTL          & 70.12          & 39.91 & 46.80          & 52.28 \\
 & TinyTL       & 78.04          & 48.12 & 48.28          & 58.15 \\
 & SparseUpdate & \textbf{81.21} & 47.83 & 50.13          & 59.72 \\ \cline{2-6} 
\multirow{-5}{*}{ProxylessNAS} &
  \cellcolor[HTML]{FFF2DF}MicroT (Ours) &
  \cellcolor[HTML]{FFF2DF}80.33 &
  \cellcolor[HTML]{FFF2DF}\textbf{51.51} &
  \cellcolor[HTML]{FFF2DF}\textbf{53.45} &
  \cellcolor[HTML]{FFF2DF}\textbf{61.76} \\ \hline
\end{tabular}
\label{tab:sota}
\end{table}

\begin{figure}[!t]
\begin{center}
\centerline{\includegraphics[width=0.45\textwidth]{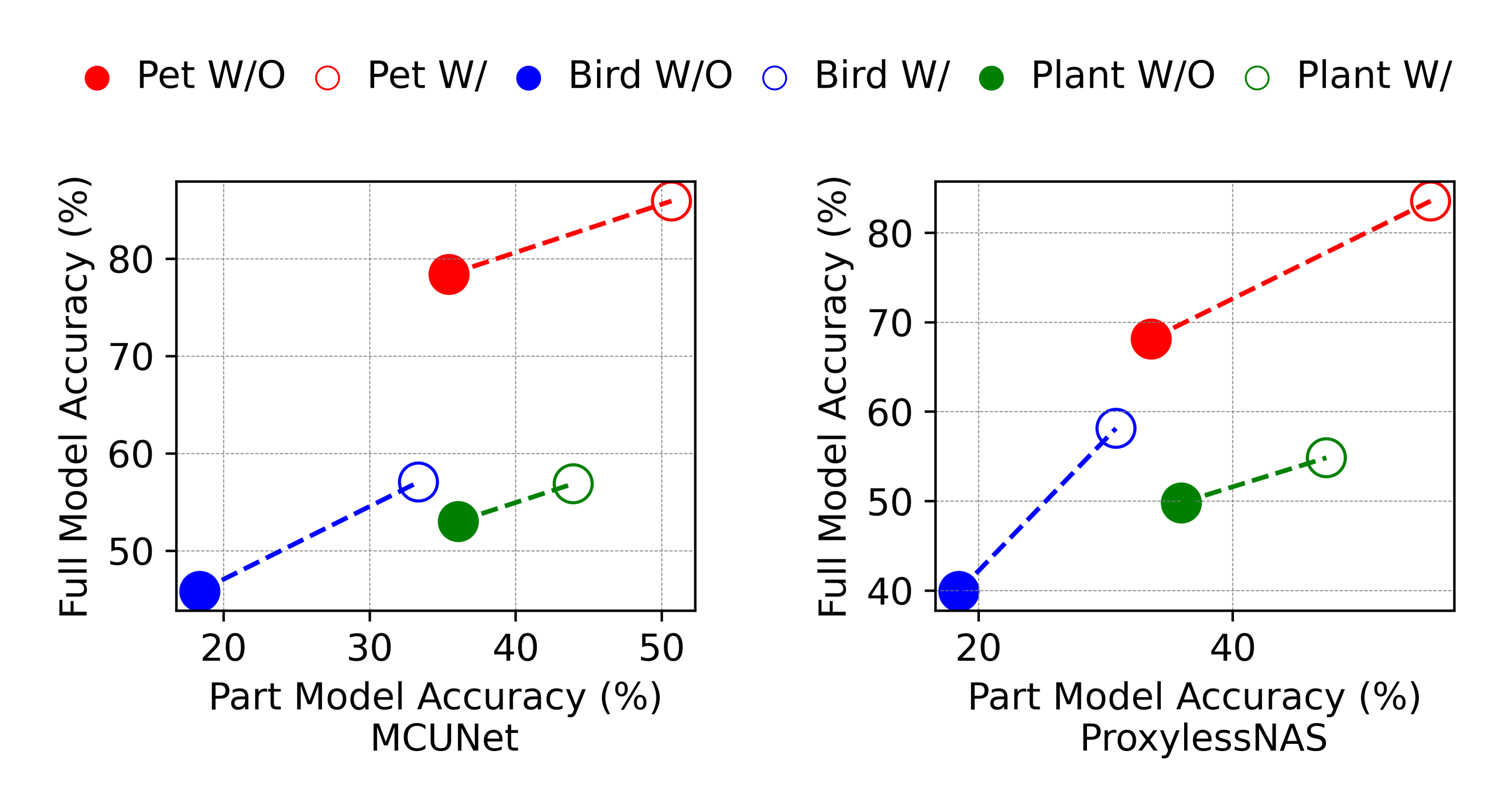}}
\caption{The performance improvement by SSKD joint training to the part/full models. The dashed connecting lines show the changes in model performance.}
\label{fig:SSKDimp}
\end{center}
\vspace{-30pt}
\end{figure}

Consequently, we split the MCUNet at its 9th module and the ProxylessNAS at its 14th module, resulting in our part/full models. These models are then further jointly trained on ImageNet using SSKD. The performance of each model With (W/) and Without (W/O) joint SSKD training is shown in Fig. ~\ref{fig:SSKDimp}. We observe that joint SSKD training part/full model performance to varying degrees. We also notice that the improvement in accuracy of the full model is slightly less than that of the part model. We believe this is the result of a combination of several factors: (i) In joint SSKD, the addition of the model segmentation point and part model could result in a degree of information loss. (ii) The accuracy improvement of the part model is also beneficial for the full model, as it can feed back and propagate to the full model. Nevertheless, the information loss in the full model is acceptable considering the performance improvements of the part/full models, and the energy saving during local joint inference (details in Section \ref{sec:energy}).

We set $MicroT$ to the standard stage-decision rate of 0.5 in order for easier compared to our the baselines, the results are shown in Table. \ref{tab:sota}. It can be observed that $None$ has the lowest accuracy across all experiments. This indicates that when deploying models locally on MCUs, personalization is necessary. Compared to $None$, $TTL$ shows significant improvement. This indicates that the transfer learning paradigm is an effective method for solving local personalization problems. Compared to $TinyTL$ and $SparseUpdate$, $TTL$ still leaves room for improvement, which suggests that merely training the classifier is insufficient to fully adapt towards the local personalized data. However, we do not abandon the 'training only the classifier' paradigm, as it offers the lowest computational requirements and energy consumption for MCUs (detailed in Section. \ref{sec:energy}). Instead, to better utilize the cloud resource available, we look towards improving the quality of the feature extractor.

$MicroT$ achieves the highest accuracy on the majority of the evaluated datasets and models, with the highest overall average accuracy. $MicroT$ outperforms $TTL$, $TinyTL$, and $SparseUpdate$ by 7.01 - 11.60 percentage points (pp), 2.29 - 5.73 pp, and 2.12 - 3.68 pp respectively. This result indicates that $MicroT$ demonstrates sufficient model performance for on-device personalization on MCUs. In addition, $MicroT$'s greatest advantage is that its reduced energy consumption (detailed in Section. \ref{sec:energy}), alongside improved performance, compared with $TTL$, $TinyTL$, and $SparseUpdate$. We do not require a complex parameter search ($SparseUpdate$) or numerous additional lite-residual modules ($TinyTL$). We achieve performance comparable to or better than these methods using a simple single-layer classifier.

\begin{table*}[!t]
\caption{The median value, stage-decision ratio, and model accuracy with different number of samples. $N$ represents the number of samples used to calculate the median value. $Med.$ represents the median value, $Ratio$ the stage-decision ratio, the $Acc.$ the model accuracy ($\%)$.}
\footnotesize
\begin{tabular}{cccccccccc|ccccccccc}
\hline
\multirow{3}{*}{N} & \multicolumn{9}{c|}{MCUNet}                                  & \multicolumn{9}{c}{ProxylessNAS}                            \\ \cline{2-19} 
 & \multicolumn{3}{c}{Pet} & \multicolumn{3}{c}{Bird} & \multicolumn{3}{c|}{Plant} & \multicolumn{3}{c}{Pet} & \multicolumn{3}{c}{Bird} & \multicolumn{3}{c}{Plant} \\ \cline{2-19} 
                   & Med.  & Ratio & Acc.  & Med.  & Ratio & Acc.  & Med.  & Ratio & Acc.  & Med.  & Ratio & Acc.  & Med.  & Ratio & Acc.  & Med.  & Ratio & Acc.  \\ \hline
5                  & 0.688 & 0.486 & 81.35 & 0.408 & 0.456 & 53.35 & 0.467 & 0.447 & 52.72 & 0.558 & 0.495 & 79.97 & 0.389 & 0.471 & 49.99 & 0.448 & 0.471 & 52.89 \\
10                 & 0.674 & 0.504 & 81.12 & 0.400   & 0.465 & 52.45 & 0.439 & 0.484 & 52.15 & 0.560 & 0.493 & 79.34 & 0.378 & 0.489 & 50.84  & 0.440 & 0.481  & 51.95 \\
20                 & 0.676 & 0.501 & 81.17 & 0.384 & 0.491 & 52.76 & 0.439 & 0.484 & 52.15 & 0.552 & 0.503 & 79.79 & 0.374 & 0.495 & 51.49 & 0.432  & 0.496 & 53.29 \\
40                 & 0.679 & 0.497 & 81.26 & 0.386 & 0.488 & 52.74 & 0.436 & 0.491 & 52.49 & 0.557  & 0.495 & 79.88 & 0.369 & 0.502 & 51.35 & 0.429 & 0.499 & 53.38 \\
All                & 0.677 & 0.500 & 81.21 & 0.378 & 0.500 & 52.85 & 0.430  & 0.500 & 52.92 & 0.554 & 0.500 & 80.33 & 0.371 & 0.500 & 51.51 & 0.428 & 0.500 & 53.45 \\ \hline
\end{tabular}
\label{tab:ratio}
\vspace{-5pt}
\end{table*}

\subsection{How Many Samples Do We Need to Determine the Confidence Thresholds?}
\label{sec:sample}

The confidence threshold dictates the stage-decision ratio, determining the proportion of samples processed only by the part model. The threshold setting is based on the confidence score sequence from the part model. Specifically, the trained part model processes $N$ samples and generates a confidence score sequence. The median of this sequence is then selected to obtain the standard stage-decision ratio of 0.5. Ideally, selecting the median of the entire test samples' confidence sequence sets the stage-decision ratio to 0.5, meaning 50\% of samples are processed only by the part model. However, in real-world applications, it is unreasonable to waste too much time finding this threshold. Therefore, to understand if it is efficient to determine the threshold locally, we randomly select various quantities of samples (regardless of category) from the entire dataset to calculate medians, and compare these medians' corresponding stage-decision ratios and model performance. We do this to understand the number of local samples needed to accurately determine the median and standard stage-decision ratio. The results are shown in Table. ~\ref{tab:ratio}.

As Table. ~\ref{tab:ratio} shows, we find that as the number of samples increases, the median, stage-decision ratio, and model performance increasingly approximate those obtained using all samples. However, even when using only five samples, the difference is minor and acceptable, with an average difference of 0.020 in median and 0.55\% in accuracy. This means processing just five samples on our MCUs is sufficient to quickly and accurately determine the standard stage-decision ratio.

\begin{figure}[!t]
\begin{center}
\centerline{\includegraphics[width=0.5\textwidth]{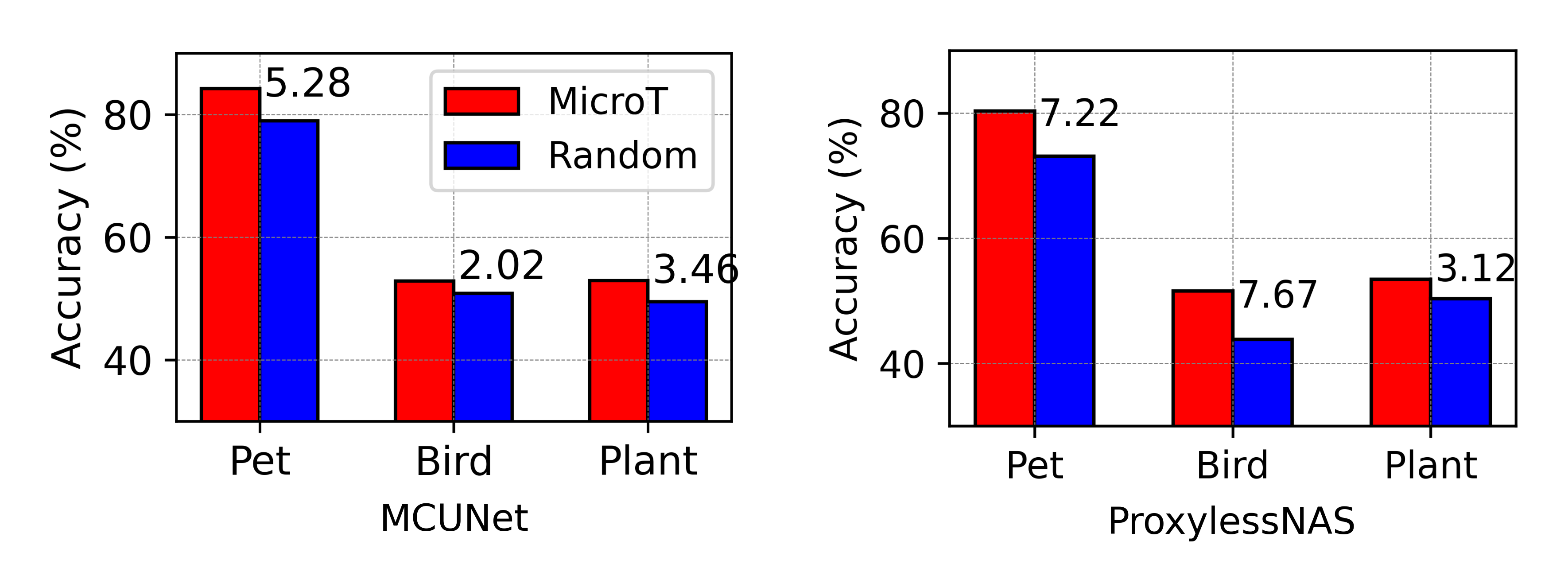}}
\caption{Comparison between stage-decision ratio determined by median or random selection.} 
\label{fig:randomratio}
\vspace{-20pt}
\end{center}
\end{figure}

Randomly selecting 50\% samples is another a way to achieve the 0.5 stage-decision ratio. Therefore, using this random selection method as our baseline, we perform a comparative experiment. As Fig. ~\ref{fig:randomratio} shows, compared to a strategy based on random selection, $MicroT$ improves the average accuracy by 4.80\%.

\section{SYSTEM COST}
\label{sec:energy}
$MicroT$ has two phases on the MCU: classifier training and inference. For the training energy cost, we compare $MicroT$ with $TTL$ and $TinyTL$. We do not compare $SparseUpdate$ as it takes \~10 days to search for the layers and channels to update and train the model, and this is inhibitively energy-consuming.  For the energy cost of inference, we select $TTL$ as our baseline ($i.e.$, full model inference), representing the inference approach used in most on-device personalization studies. We select the Plant \cite{goeau2017plant} and Bird datasets \cite{WahCUB_200_2011}, as these datasets contain the smallest and largest number of classes respectively. To efficiently analyze the energy cost, we use the average of these two datasets.

We measure the memory usage of $MicroT$ using the STM32-CUBE-IDE \cite{cubeide}. To effectively show that $MicroT$ can meet the memory requirements: (i) We only analyze the local training stage, since the memory usage of the local training stage is higher than the inference stage. (ii) We only analyze the full model, since the memory usage of the full model is higher than that of the part model. The maximum RAM usage of MCUNet, at 224 resolution, is 614.40$KB$ and 624.64$KB$ for ProxylessNAS. The flash memory usage of MCUNet is \~0.91$MB$ and ~0.70$MB$ for ProxylessNAS. These results demonstrate $MicroT$ can meet the memory constraints of commercial MCUs.

\begin{figure}[!t]
\begin{center}
\centerline{\includegraphics[width=0.5\textwidth]{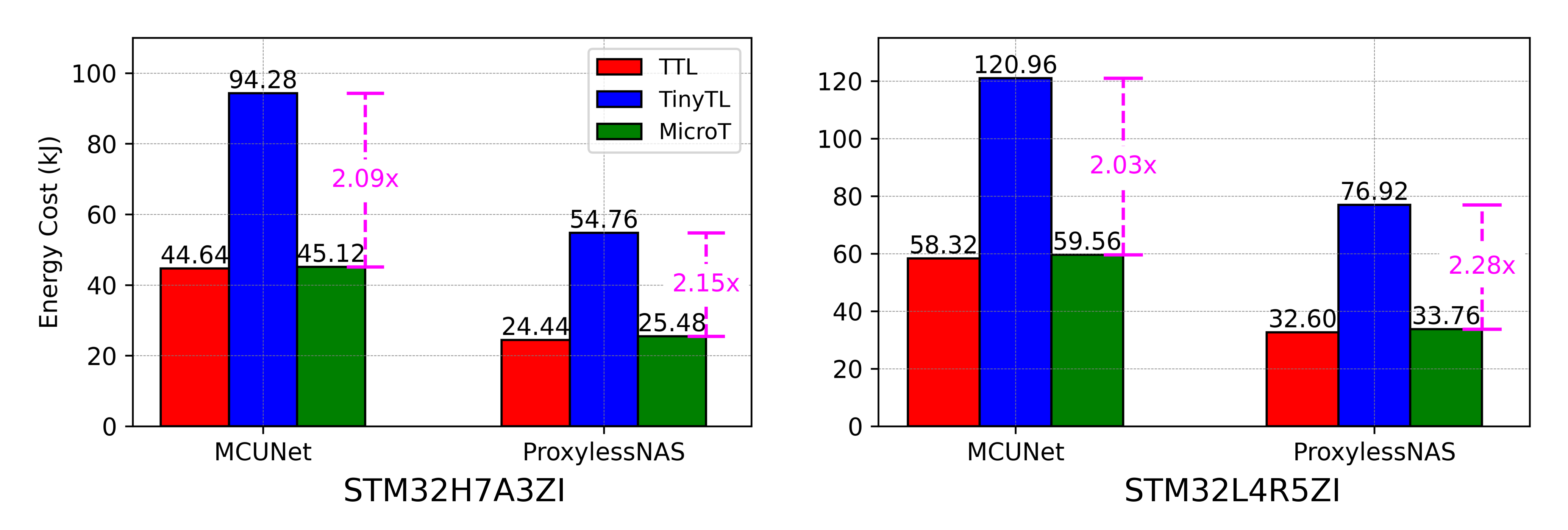}}
\caption{Energy Cost of Local Training.}
\label{fig:energycost}
\end{center}
\vspace{-25pt}
\end{figure}

We utilize the Monsoon Power Monitor \cite{msoon} to measure the time and average power of training, then calculate their energy consumption (as described in Section. \ref{sec: metrics}).  To facilitate comparison and measurement of the energy consumption, we use the example of training 100 local data samples for 40 epochs ($i.e.$, 4,000 iterations total). We ignore the energy consumption of other MCU components or operations and only consider the energy consumption of local training, including data preprocessing, feature extractor inference, and classifier training. The results are shown in Fig. ~\ref{fig:energycost}.

It can be observed that during the local training, the energy consumption of $MicroT$ is not significantly different from that of $TTL$. Although $MicroT$ needs to train two classifiers for the part/full models, we employ a stage-training strategy to save energy: the sample first enters the part model to train its classifier, saving the intermediate values from the layer preceding the part model classifier. This enables us to continue using the pre-calculated intermediate values when training the full model classifier. Therefore, the difference between $MicroT$ and $TTL$ mainly lies in the training of the part model classifier and the cost of storing/transmitting a small number of parameters. Moreover, our classifier structure is simple. The aforementioned factors result in similar energy consumption between $MicroT$ and $TTL$. In contrast, $TinyTL$ consumes more energy because it requires training numerous and complex lite-residual modules. Compared to $TinyTL$, $MicroT$ saves 2.03 to 2.28$\times$ the energy. This improvement is attributed to $MicroT$'s simple model architecture and efficient optimization strategy during local training.

\begin{table}[!t]
\caption{The energy costs ($kJ$) of the MCU local inference. MicroT achieves energy saving on two boards and two models.}
\footnotesize
\begin{tabular}{
>{\columncolor[HTML]{FFFFFF}}c 
>{\columncolor[HTML]{FFFFFF}}c 
>{\columncolor[HTML]{FFFFFF}}c 
>{\columncolor[HTML]{FFFFFF}}c 
>{\columncolor[HTML]{FFFFFF}}c }
\hline
\textbf{Model}           & \textbf{Method}            & \textbf{STM32H7A3ZI}        & \textbf{STM32L4R5ZI}        & {\color[HTML]{000000} \textbf{Avg.}} \\ \hline
\cellcolor[HTML]{FFFFFF} & TTL                        & 9.52                        & 12.42                       & {\color[HTML]{000000} 10.97}         \\
\multirow{-2}{*}{\cellcolor[HTML]{FFFFFF}MCUNet} &
  {\color[HTML]{000000} MicroT} &
  {\color[HTML]{000000} \textbf{8.22}} &
  {\color[HTML]{000000} \textbf{10.54}} &
  {\color[HTML]{000000} \textbf{9.38}} \\ \hline
\cellcolor[HTML]{FFFFFF} & {\color[HTML]{000000} TTL} & {\color[HTML]{000000} 4.64} & {\color[HTML]{000000} 6.19} & {\color[HTML]{000000} 5.42}          \\
\multirow{-2}{*}{\cellcolor[HTML]{FFFFFF}ProxylessNAS} &
  {\color[HTML]{000000} MicroT} &
  {\color[HTML]{000000} \textbf{4.01}} &
  {\color[HTML]{000000} \textbf{5.31}} &
  {\color[HTML]{000000} \textbf{4.66}} \\ \hline
\end{tabular}
\label{tab:infenergy}
\vspace{-5pt}
\end{table}

Next, we evaluate the inference energy cost. We choose $TTL$ ($i.e.$, full model inference) as our baseline as it is used in most current research. We only focus on the energy cost during model inference, ignoring the energy consumption of other MCU components. We have the MCU perform 1,000 model inferences. The results are shown in Table.~\ref{tab:infenergy}.

We first observed that $MicroT$ achieves a reduction in energy consumption during model inference across both boards and models. Compared to $TTL$, $MicroT$ reduces energy consumption by an average of 14.17\%, potentially saving a significant amount of energy in the long run. We also observe that the energy consumption of MCUNet than ProxylessNAS is higher, due to the more complex computations involved in MCUNet. Additionally, for the same model, the energy consumption of the STM32L4R5ZI MCU is higher than that of the STM32H7A3ZI, despite the STM32L4R5ZI being lower power. This is due to increased running time, highlighting the trade-off between running time and power for energy consumption.

\begin{table}[!t]
\caption{The model performances (\%) with various stage-decision ratios.}
\footnotesize
\label{tab:performanceratio}
\begin{tabular}{cccccc}
\hline
\textbf{Model}& \textbf{Method} &\textbf{Pet} & \textbf{Bird}  & \textbf{Plant} & \textbf{Avg.}  \\ \hline
\multirow{2}{*}{MCUNet}       & MicroT\_0.25 & 84.34 & 56.05 & 53.87 & 64.75 \\
                              & MicroT\_0.75 & 76.68 & 47.87 & 51.65 & 58.73 \\ \hline
\multirow{2}{*}{ProxylessNAS} & MicroT\_0.25 & 82.35 & 51.97 & 54.04 & 62.79 \\
                              & MicroT\_0.75 & 76.44 & 46.59 & 50.36 & 57.80 \\ \hline
\end{tabular}
\vspace{-5pt}
\end{table}

\begin{table}[!t]
\caption{The energy costs ($kJ$) of the MCU local inference with various stage-decision ratios.}
\footnotesize
\begin{tabular}{ccccc}
\hline
\textbf{Model} & \textbf{Method} & \textbf{STM32H7A3ZI} & \textbf{STM32L4R5ZI} & \textbf{Avg.} \\ \hline
\multirow{3}{*}{MCUNet}       & TTL          & 9.52 & 12.42 & 10.97 \\
                              & MicroT\_0.25 & 8.67 & 11.28 & 9.98  \\
                              & MicroT\_0.75 & 7.75 & 9.61  & 8.68  \\ \hline
\multirow{3}{*}{Proxy} & TTL          & 4.64 & 6.19  & 5.42  \\
                              & MicroT\_0.25 & 4.23 & 5.65  & 4.94  \\
                              & MicroT\_0.75 & 3.71 & 4.87  & 4.29  \\ \hline
\end{tabular}
\label{tab:energyratio}
\vspace{-5pt}
\end{table}

\begin{table}[!t]
\caption{The impact of quantization on accuracy (\%).}
\footnotesize
\begin{tabular}{cccclc}
\hline
\textbf{Model}                & \textbf{Quant.} & \textbf{Pet} & \textbf{Bird} & \textbf{Bird} & \textbf{Avg.} \\ \hline
MCUNet & W/O & 81.64 & 53.13 & 53.19 & 62.65 \\
       & W/  & 81.21 & 52.85 & 52.92 & 62.33 \\ \hline
\multirow{2}{*}{ProxylessNAS} & W/O             & 80.82        & 52.21         & 54.26         & 62.43         \\
       & W   & 80.33 & 51.51 & 53.45 & 61.76 \\ \hline
\end{tabular}
\label{tab:quant}
\vspace{-5pt}
\end{table}

\begin{figure}[!t]
\begin{center}
\centerline{\includegraphics[width=0.5\textwidth]{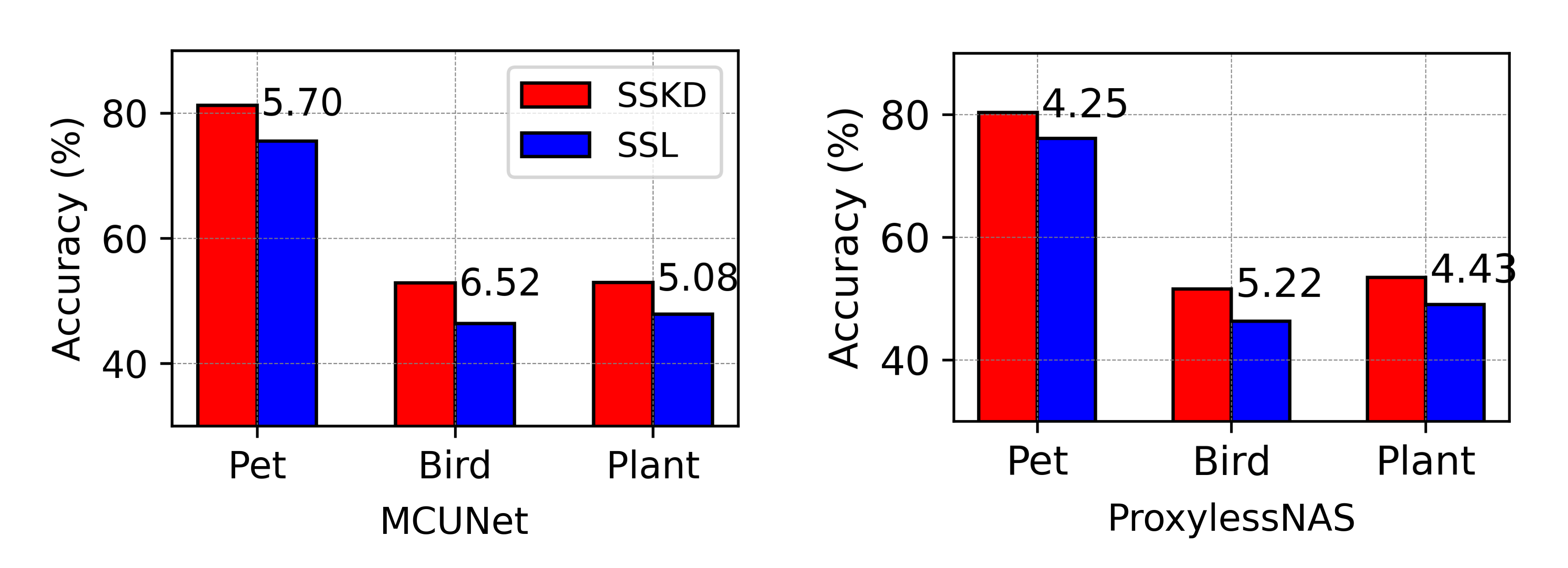}}
\caption{The performance between SSL and SSKD to the feature extractor.}
\label{fig:ablationSSL}
\vspace{-20pt}
\end{center}
\end{figure}

\section{ABLATION STUDY}
\label{sec:ablation}

\textbf{SSL vesus SSKD.} We evaluated the difference in model performance between directly using SSL for the feature extractor and using SSKD, with the results shown in the Fig. \ref{fig:ablationSSL}. We can observe that directly applying SSL to the feature extractor results in less improvement in model performance compared to using SSKD. This is mainly because the models used on the MCU are relatively small, and the performance optimization achieved by directly applying SSL to these models is limited by their capacity. In contrast, applying SSL to a large model (teacher model) first and then having the small model (student model) learn both the true labels and the teacher's embedded features makes it easier for the small model to acquire this knowledge.

\textbf{Model performance with various stage-decision ratio.} In Section.~\ref{sec:performance}, we have introduced the model performance of $MicroT$ with ratio of 0.5. To provide a clear demonstration of the performance trend, we further show the model performance with ratios of 0.25 and 0.75. The results are shown in Table.~\ref{tab:performanceratio}.

\textbf{Energy saving with various stage-decision ratio.} We evaluated the energy cost of local inference at different stage-decision ratios, including ratios of 0.25 and 0.75. A ratio of 0 or 1 means that the inference is performed entirely by the full or part model, and we did not consider these extreme cases. We still compared $MicroT$ with $TTL$. The results are shown in the Table. \ref{tab:energyratio}. We can observe that at different stage-decision ratios, $MicroT$ can achieve varying degrees of energy savings. Compared to $TTL$, $MicroT\_0.25$ saved an average of 8.94\% energy, while $MicroT\_0.75$ saved an average of 20.81\%. This demonstrates that the configurable stage-decision ratio of $MicroT$ provides users with a flexible configuration to balance model performance and energy cost.

\textbf{The impact of quantization.} MicroT uses partial quantization, where the feature extractor is quantized to \texttt{INT8} while the classifier created locally on the MCU retains the \texttt{FLOAT32}. This approach allows the feature extractor, which accounts for a larger proportion of parameters, to reduce memory usage, improve inference speed, and decrease energy consumption. At the same time, the locally trained classifier can compensate for the accuracy loss caused by the quantized feature extractor through training. The results W/ and W/O partial quantization are compared in Table. \ref{tab:quant}. We can observe that the average accuracy difference is approximately 0.5\%.



\section{DISCUSSION}
\label{sec:discussion}


In this section, we discuss some analysis and limitations of the methodology and experiments in this study, as well as the future works.

(1) \textbf{Application scenarios.} In this study, we primarily conducted experiments in the image domain, which offers many open-source models/datasets, making it convenient for us to validate $MicroT$. However, $MicroT$ has several application scenarios in energy-sensitive environments. For instance, battery-free underwater devices powered by acoustic energy that use low-power cameras to detect and identify marine species \cite{afzal2022battery}; solar-powered low-power sensor networks for monitoring wildlife \cite{hart2020precision}; environmental monitoring devices powered by wind or water energy that require long-term self-sustainability \cite{samijayani2023solar}; and implantable medical devices powered by bioenergy \cite{jiang2020emerging}. These practical applications can all leverage the technologies related to $MicroT$.

(2) \textbf{The experimental models and MCUs.} We have currently tested the performance of $MicroT$ on a limited number of models, but its versatility enables it to be easily applied to other lightweight models, such as SqueezeNet \cite{iandola2016squeezenet}, MobileNet \cite{howard2017mobilenets}, and models generated by Neural Architecture Search (NAS). $MicroT$ can also be easily combined with other model compression methods, such as quantization \cite{xiao2023smoothquant,jacob2018quantization} and pruning \cite{zhu2017prune,liu2020pruning}.

(3) \textbf{Stability of model segmentation.} One of the objectives of the feature extractor is to enhance its generalization, which means that the feature extractor can also exhibit high generalization on local datasets. Therefore, the global optimal segmentation point is potentially to be the optimal or near-optimal point on the local dataset as well. However, if the local optimal point differs from the global point, our strategy is to conservatively adopt the global point because it is a more stable and general choice. Even if the optimal point varies slightly on the local dataset, the overall performance could still meet requirements. 

(4) \textbf{The `feature extractor + classifier' paradigm.} This paradigm has advantages in energy efficiency, which is the core focus of $MicroT$ for ultra-low-power devices. However, to reduce resource consumption, local training only involves the classifier, which places high demands on the generalization of the feature extractor. Nevertheless, this is a necessary design consideration on performance and energy efficiency. $MicroT$ remains effective in most real-world application scenarios because the generalization of the feature extractor can be ensured: first, the rapid development of smart devices has provided a wealth of publicly available training datasets; second, we can leverage technologies such as SSL.

(5) \textbf{Re-training classifiers.} Ideally, once the classifier is trained locally and meets task requirements, it can stop training and be used for a long time. However, in practice, local data often undergoes some degree of drift, such as changes in lighting conditions due to seasonal variations. While the general feature extractor can handle minor local drifts, in some extreme cases, further fine-tuning of the classifier may still be necessary. We believe that this fine-tuning is usually quick and straightforward, as local data distribution changes are often gradual rather than sudden.

(6) \textbf{Unsupervised Personalization.} $MicroT$ relies on labeled data for local training, similar to most on-device personalization approaches \cite{cai2020tinytl,lin2022device,kwon2023tinytrain}. However, exploring unsupervised training, such as distance-based k-means classifiers \cite{wu2020emo}, is an interesting direction. We conducted experiments with traditional k-means classifiers and found they performed well on simple datasets. However, the traditional k-means classifiers did not perform well on complex datasets, likely due to the large number of categories and dense feature space. Nevertheless, improving the differentiation of categories in the feature space through methods such as contrastive learning \cite{chen2020simple} may enhance the performance of unsupervised personalization. Such advancements would broaden the applicability of $MicroT$.

(7) \textbf{Multi-modal Feature.} We only evaluated $MicroT$ using image data as an example. However, the ability to manage diverse signal sources, like environmental sensors \cite{daghero2022human,kaewmard2014sensor}, is also essential for MCUs. Recent studies have investigated data and feature fusion techniques for multi-modal data analysis on edge devices \cite{wang2022am3net,ouyang2022cosmo}. $MicroT$ could integrate its feature extractor with these fusion methods to map multi-modal data into a unified feature space. This expansion not only widens $MicroT$'s utility but also enhances its robustness. Other tasks, such as person detection, vehicle counting, and audio classification, are also interesting future research directions and applications.

\section{CONCLUSION}
\label{sec:conclusion}


Current approaches for ML model personalization on MCUs either rely on communication with cloud devices or require complex pre-training/training processes on the MCU. Both of these processes consume significant amounts of energy, undermining the low-energy advantage of MCUs. To address this, we propose $MicroT$, an efficient framework that uses self-supervised knowledge distillation alongside an early-exit mechanism. $MicroT$ outperforms traditional transfer learning (TTL) and two SOTA approaches by 7.01 - 11.60\%, 2.29 - 5.73\%, and 2.12 - 3.68\% in accuracy, respectively. During the MCU training phase, $MicroT$ saves 2.03 - 2.28$\times$ energy compared to SOTA methods. During the MCU inference phase, $MicroT$ reduces energy consumption by 14.17\% compared to $TTL$.

\newpage

\bibliographystyle{unsrt}
\bibliography{sample-base}











\end{document}